\documentclass[conference]{IEEEtran}
\pdfoutput=1
\usepackage{cite}
\usepackage{amsmath,amssymb,amsfonts}
\usepackage{algorithmic}
\usepackage{graphicx}
\usepackage{textcomp}
\usepackage{xcolor}
\usepackage{multirow}
\def\BibTeX{{\rm B\kern-.05em{\sc i\kern-.025em b}\kern-.08em
    T\kern-.1667em\lower.7ex\hbox{E}\kern-.125emX}}
\begin{document}

\title{FPGA Implementation of Minimum Mean Brightness Error Bi-Histogram Equalization\\}

\makeatletter
\newcommand{\linebreakand}{%
  \end{@IEEEauthorhalign}
  \hfill\mbox{}\par
  \mbox{}\hfill\begin{@IEEEauthorhalign}
}
\makeatother

\author{
\IEEEauthorblockN{Abhishek Saroha}
\IEEEauthorblockA{\textit{Dept. of Computer Science and Engineering} \\
\textit{National Institute of Technology}\\
Delhi, India
}
\and
\IEEEauthorblockN{Avichal Rakesh}
\IEEEauthorblockA{\textit{Department of Computer Science} \\
\textit{University of Wisconsin-Madison}\\
Madison, USA \\
}
\linebreakand
\IEEEauthorblockN{Rajiv Kumar Tripathi}
\IEEEauthorblockA{\textit{Dept. of Electronics and Communication Engineering} \\
\textit{National Institute of Technology}\\
Delhi, India \\
}
}

\maketitle

\begin{abstract}
Histogram Equalization (HE) is a popular method for contrast enhancement. Generally, mean brightness is not conserved in Histogram Equalization. Initially, Bi-Histogram Equalization (BBHE) was proposed to enhance contrast while maintaining a the mean brightness. However, when mean brightness is primary concern, Minimum Mean Brightness Error Bi-Histogram Equalization (MMBEBHE) is the best technique. There are several implementations of Histogram Equalization on FPGA, however to our knowledge MMBEBHE has not been implemented on FPGAs before. Therefore, we present an implementation of MMBEBHE on FPGA.
\end{abstract}

\begin{IEEEkeywords}
Histogram Equalization, Field Programmable Gate Array, Image Enhancement
\end{IEEEkeywords}

\section{Introduction}
Histogram Equalization is a popular image processing technique for image enhancement. It uses the image frequency histogram to change the image's contrast. This often results in a change in image brightness, which can lead to visual artifacts not present in the original image \cite{b1, b2, b3}. However, there are certain cases (for example, consumer electronic products such as TV) where brightness of an image needs to be preserved to a larger degree. Bi-Histogram Equalization \cite{b4} was proposed to conserve mean brightness along with justified contrast enhancement. But, for maximum mean brightness conservation Minimum Mean Brightness Error Bi-Histogram Equalization \cite{b5} was proposed. MMBEBHE ensures that the mean brightness of the output image is as close to the original as possible with contrast enhancement. This makes MMBEBHE computationally more expensive than HE and BBHE, and thus, makes it difficult to implement in a constrained environment.

A Field programmable gate array (FPGA) is an integrated circuit, made of a number of programmable logic gates, which can vary from tens of thousands to millions. These logic gates have interconnections programmable by the user. This makes FPGAs extremely useful for a variety of applications, both academic, and corporate. They are most commonly used to create accelerated hardware where execution of an algorithm is optimized by the hardware. This is often the first step to creating specialized hardware. Initial work on FPGA implementation of image processing has been carried out by Trost et al. in \cite{b6}. FPGA implementation of contrast enhancement of an image is proposed \cite{b7} and specifically for HE has been proposed in \cite{b8,b9}. Implementation of MMBEBHE on FPGA is proposed in this paper.

\section{System Description}
We use a Xilinx Basys3 board for our implementation. The Basys3 is based on Artix-7. It is a starter board of relatively low cost, and has VGA, USB, along with other ports. It’s features make it suitable for a variety of different circuits.

Initial testing, validation and timings were done on ModelSim, and later migrated to Xilinx's Vivado Design Suite for more accurate simulations and timing. The results shown in this paper are from Vivado Design Suite's simulations.

\section{MMBEBHE and Modifications}
It is difficult and expensive to work with floating point numbers on FPGAs, so we modify the MMBEBHE to use integer arithmetic only. This section details the mathematical modifications made to the original description of MMBEBHE to limit operations to integers.

\textit{Histogram Equalization} defines the the probability density function $P(X_{k})$ as
\begin{equation*}
    P(X_{k}) = \frac{n^{k}}{n}
\end{equation*}

Where $n$ is the total number of pixels in $\textbf{X}$, and $n^{k}$ is the number of times $X_{k}$ appears in $\textbf{X}$, and $k = 0, 1, 2, \ldots, L - 1$ 

Notice that $n$ is constant across all values of $P(X_{k})$. So, instead of storing floating point $P(X_{k})$, we can store $n^{k}$ and $n$ separately. 

Next, \textit{Histogram Equalization} defines cumulative density function $c(k)$ as
\begin{equation}
    c(k) = \sum^{k}_{j = 0}P(X_{k}) \label{eq1}
\end{equation}
$c(k)$ as defined by HE deals with floating point numbers. We thus make the following calculations
\begin{align}
    c(k) & = \sum^{k}_{j = 0}P(X_{j}) \notag\\
    & = \frac{n^{0}}{n} + \frac{n^{1}}{n} + \frac{n^{2}}{n} + \ldots + \frac{n^{k}}{n} \notag\\
    & = \frac{1}{n}(n^{0} + n^{1} + n^{2} + \ldots + n^{k}) \notag\\
    & = \frac{1}{n}\sum^{k}_{j = 0}n^{j} \notag\\
    c(k) & = \frac{1}{n} \cdot f_{c}(X_{k}) \label{eq2}
\end{align}
Where $f_{c}(X_{k})$ is the cumulative frequency of $X_{k}$. As with probability density function, values of $n$, and $f_{c}(X_{k})$ can be separately. We define $f_{c}(X_{k})$ as
\begin{align}
    f_{c}(X_{k}) = \sum_{j = 0}^{k}{n^{j}} \label{eq3}
\end{align}
For creating the map, \textit{Histogram Equalization} defines a function $F(k)$ as 
\begin{align}
    F(k) = X_{0} + (X_{L - 1} - X_{0}) \cdot c(k)
\end{align}
Using the value of $c(k)$ from \eqref{eq2}, we get
\begin{align}
    F(k) & = X_{0} + (X_{L - 1} - X_{0}) \cdot c(k) \notag\\
    & = X_{0} + (X_{L - 1} - X_{0}) \cdot \frac{1}{n} \cdot f_{c}(X_{k}) \notag\\
    F(k) & = \frac{nX_{0} + (X_{L - 1} - X_{0}) \cdot f_{c}(X_{k})}{n} \label{eq4}
\end{align}
As $n$, $X_{0}$, $X_{L - 1}$, and $f_{c}(X_{k})$ are all integers, the numerator of \eqref{eq4} can be calculated without floating point operations. Also, since $F(k)$ maps a pixel value to a pixel value, we can use integer division to calculate the map. However, integer division itself won't round up the pixel values. Rounding will need to be performed using the mod($\%$) operator. 

These changes reduce the overhead of floating point operations, without changing the mathematics behind Histogram Equalization, however, they increase the complexity of rounding. The mod($\%$) operator increases the complexity of the synthesized hardware.

The output image $\textbf{Y} = Y(i, j)$ is then calculated as
\begin{align}
    \textbf{Y} & = F(\textbf{X}) \notag\\
    & = F(X(i, j))\;|\;\forall\:X(i, j) \in \textbf{X} \label{eq5}
\end{align}

With these changes made to \textit{Histogram Equalization}, the rest of MMBEBHE is followed as normal. The \textit{Scaled Mean Brightness Error} (\textit{SMBE}) is calculated for each intensity value as

\begin{align}
    SMBE_{0} & = L(n - F(X_{0})) - 2\sum_{i = 0}^{L-1}iF(X_{i}) \label{eq6}\\
    SMBE_{\gamma} & = SMBE_{\gamma - 1} + [n - LF(X_{\gamma})] \label{eq7}
\end{align}
where $F(X_{k}) = n^k$, and $L$ is the range of possible pixel values ($256$ in our case).

The $threshold$ is calculated as the intensity values at which absolute value of $SMBE$ is the least. The histogram is split along the $threshold$, and HE is independently performed on each half.

\begin{figure*}[ht!]
    \begin{center}
        \includegraphics[width=16.0cm]{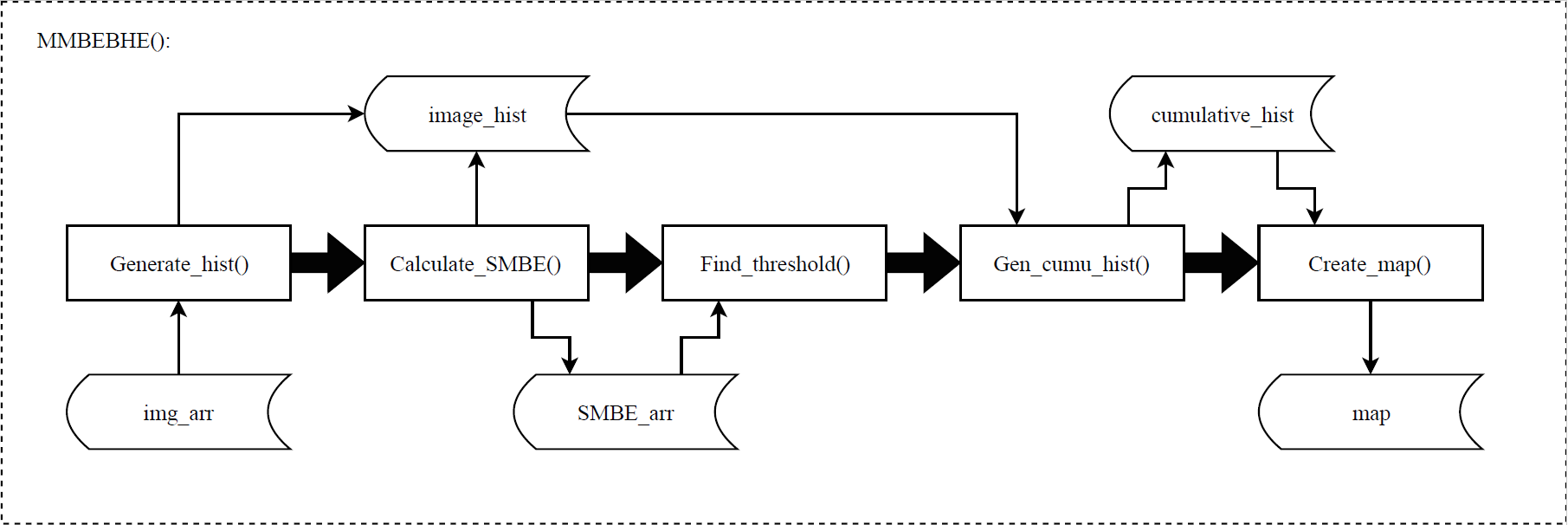}
        \caption{FPGA implementation flowchart}
        \label{fig1}
    \end{center}
\end{figure*}

\section{Implementation on FPGA}

Following are the details of how MMBEBHE was implemented on Basys 3. The algorithm was broken down into logical modules. Each module was tested separately before being pipelined for the final result. Our implementation works on 8-bit image of arbitrary size. However, due to the complexity of the generated schematic, we have included a bare-bones schematic for a binary image with 8 pixels which is easier to comprehend.

Figure~\ref{fig1} shows the high-level interaction between different modules. The execution stops after the output map is calculated. Although this paper presents the implementation as different modules, the final synthesis was coalesced into a single \texttt{mmbebhe} module, which takes image as input and outputs the corresponding map.

\subsection{Generate\_hist()}

\begin{figure*}[ht!]
    \begin{center}
        \includegraphics[width=12.0cm]{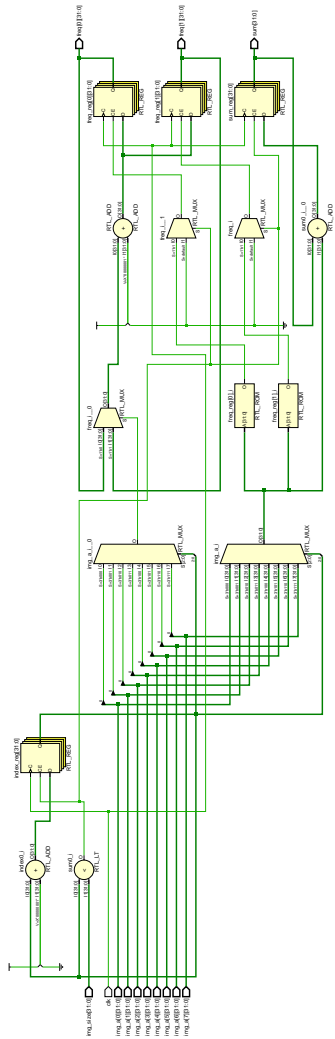}
        \caption{Generate\_histogram() schematic for binary image with 8 pixels}
        \label{fig2}
    \end{center}
\end{figure*}

This module takes the image as input, and outputs a histogram of frequency of each pixel value. Internally, this module contains a pointer to the image array and uses it to access one element per clock cycle. This retrieved value is used to increment the frequency in a histogram that is also kept internally, and added to a register sum, which tracks the sum of all the pixels seen. Once the module has iterated over all the pixels, a done flag is set to 1. The saved histogram is sent further down the pipe to the other modules that need it, and the execution within the module stops.

The input \texttt{image} array splits into two branches: one for calculating the frequency histogram \texttt{freq[255]}, and another for calculating the \texttt{sum} of all pixels in the image. For calculating \texttt{freq}, the pixel value at \texttt{index} is extracted from \texttt{image}, and \texttt{1} is added to the register corresponding to the pixel value. For calculating \texttt{sum}, pixel value is extracted, and passed through an adder along with a register, \texttt{sum}, which stores the running sum of all seen pixels.

\subsection{Calculate\_smbe()}

\begin{figure*}[ht!]
    \begin{center}
        \includegraphics[width=17.0cm]{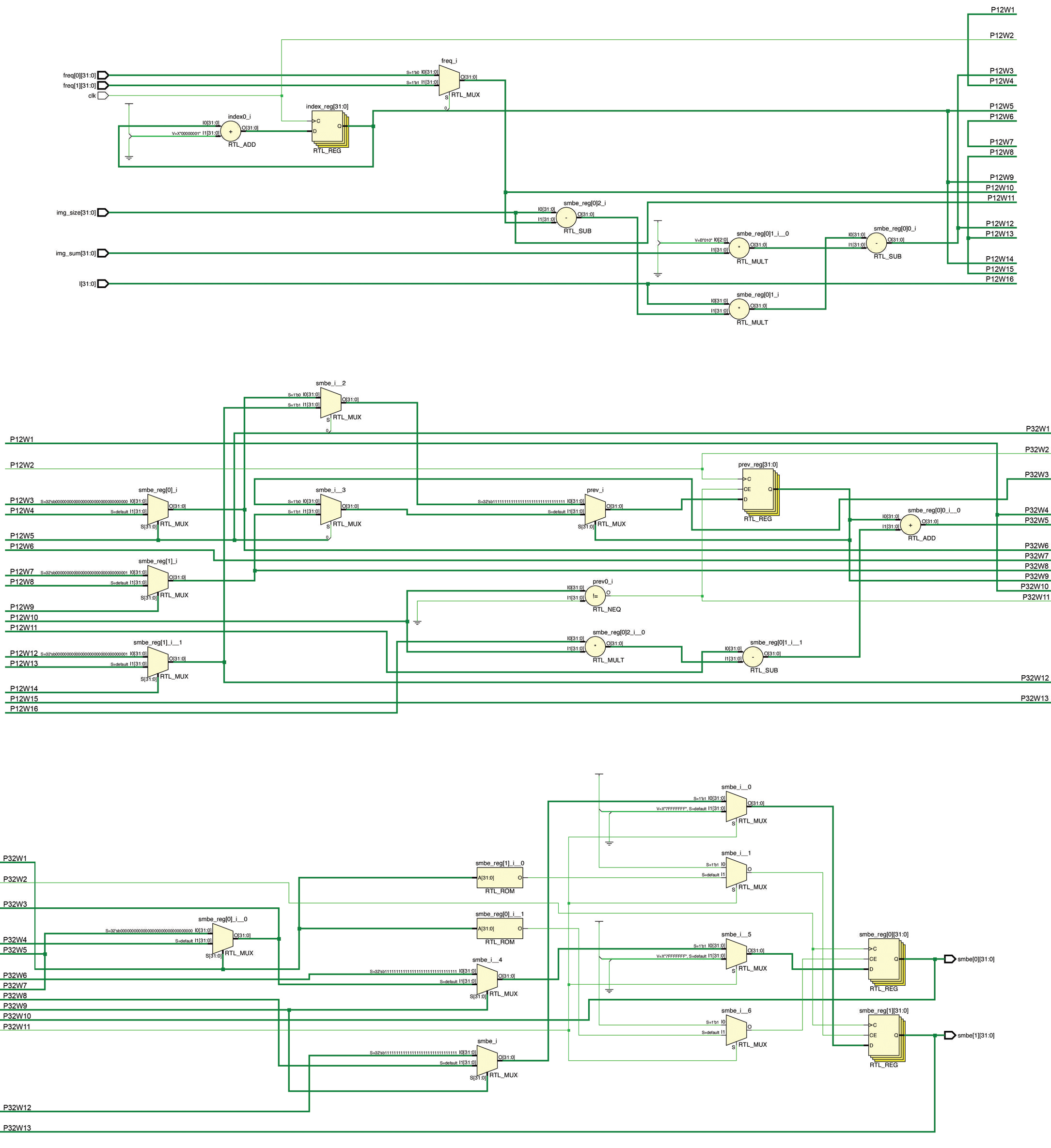}
        \caption{Calculate\_SMBE() for a binary image with 8 pixels}
        \label{fig3}
    \end{center}
\end{figure*}

This module is responsible for calculating the SMBE for each pixel value present in the image. For a pixel value not present in the image, the SMBE is set to \texttt{0x7fffffff}, to ensure it will not be chosen as the threshold. Calculate\_SMBE() takes the histogram, and sum from Generate\_Hist() as input. Internally, it iterates through the histogram, and for each pixel value calculates the SMBE as described in equation \eqref{eq6} and \eqref{eq7} and stores it in a map. Note that in equation \eqref{eq6}, $ \sum^{L - 1}_{i = 0}{i \cdot F(X_{i})} $ is the sum of all pixels in the image. This sum is calculated by Generate\_Hist() and simply consumed by this module. Once all SMBEs are calculated, a \texttt{done} flag is tripped to stop execution of the module and trigger the next step in the pipeline. 

Since calculating SMBE values is defined recursively, with each SMBE depending on the previous one, the module uses a 32-bit register \texttt{prev} to store the previous entry. A register \texttt{first} is initialized to \texttt{0}, as sentinel for the base case. The input frequency array \texttt{freq} goes through a multiplexer which uses \texttt{index} as the selector to iterate through each pixel's frequency. The selected frequency is compared to 0. If frequency is 0, the corresponding SMBE is set to \texttt{0x7fffffff}. Otherwise, the recursive formula is followed. This repeats serially for each pixel value in \texttt{freq} array, with \texttt{prev} getting updated whenever a non-zero frequency is processed. It is important to execute this part serially due to dependence on the previous element.

\subsection{Find\_Threshold()}

\begin{figure*}[ht!]
    \begin{center}
        \includegraphics[width=15.0cm]{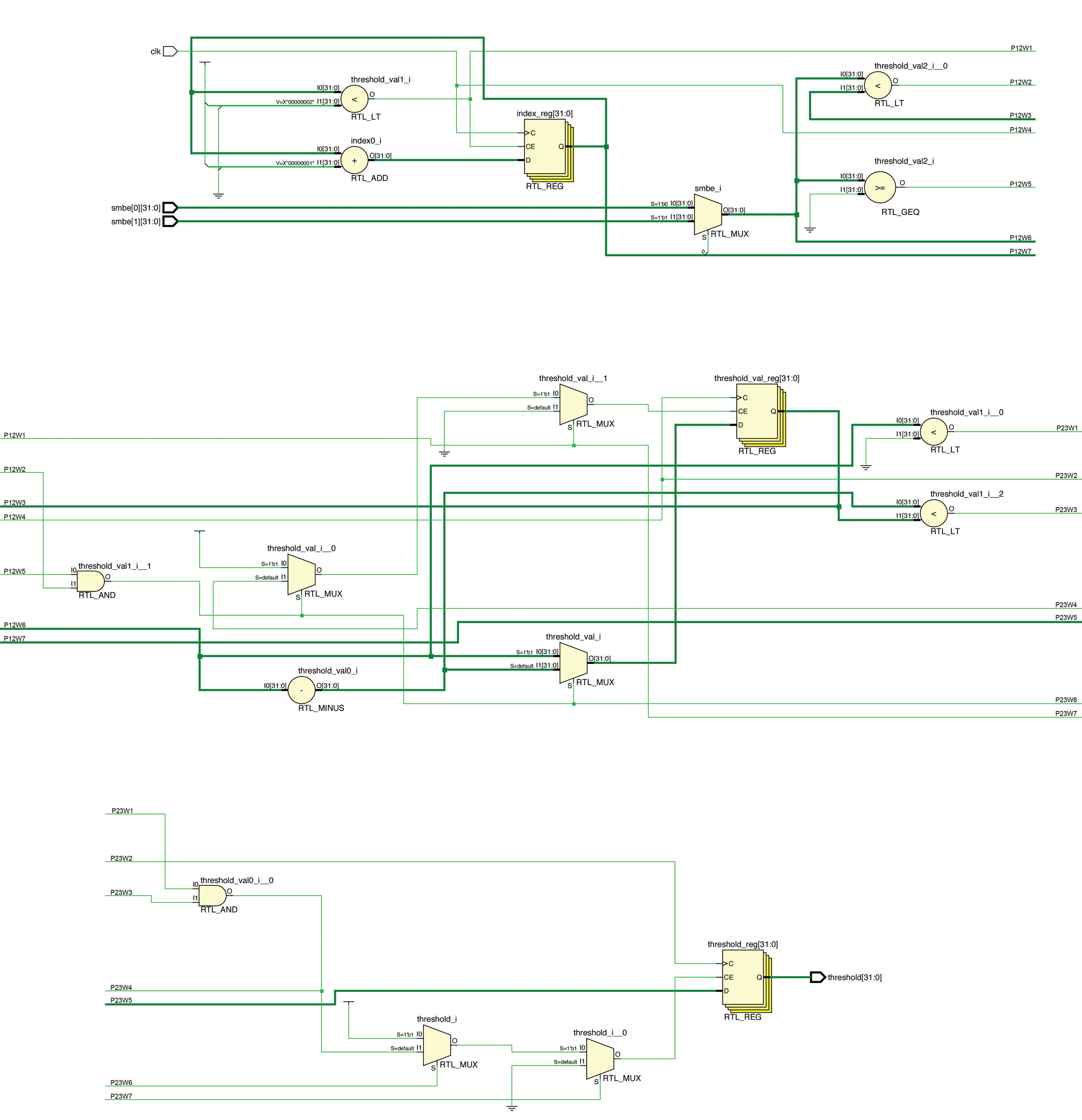}
        \caption{Find\_threshold() schematic for a binary image with 8 pixels}
        \label{fig4}
    \end{center}
\end{figure*}

This module takes the SMBE map from Calculate\_SMBE() and looks for the pixel value which has absolute minimum SMBE. This value is the threshold along which the histogram will be divided further down the pipeline. This value is stored and passed forward along with a \texttt{done} flag.

The input SMBE values pass through a multiplexer with \texttt{index} as selector. The selected \texttt{SMBE\_val} goes through comparators that compare the absolute value of \texttt{SMBE\_val} to the current \texttt{threshold\_val}. The absolute value comparison is evaluated as follows: \texttt{SMBE\_val < 0} \texttt{\&\&} \texttt{-SMBE\_val < threshold\_val}, and \texttt{SMBE\_val >= 0} \texttt{\&\&} \texttt{SMBE\_val < threshold\_val}. If either condition evaluates to true, \texttt{threshold\_val} is set to \texttt{SMBE\_val}, or \texttt{-SMBE\_val} as appropriate, and \texttt{index} is saved for output. \texttt{index} then increments by \texttt{1}, and the process repeats until \texttt{index < 256}.

\subsection{Gen\_cumu\_hist()}

\begin{figure*}[ht!]
    \begin{center}
        \includegraphics[width=17.0cm]{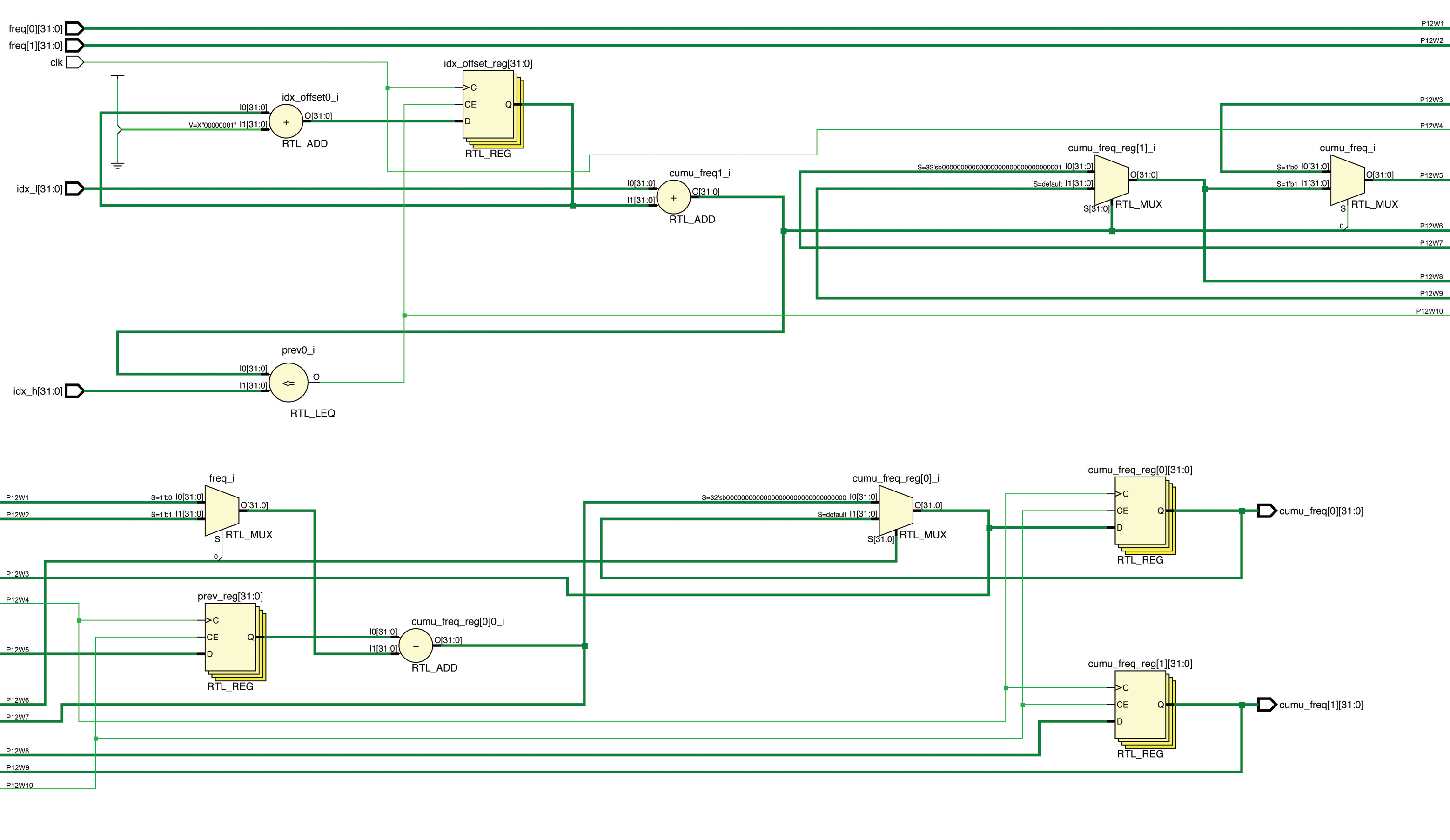}
        \caption{Gen\_cumu\_hist() schematic for a binary image with 8 pixels}
        \label{fig5}
    \end{center}
\end{figure*}

This module takes the histogram from Generate\_Hist(), a lower bound, and an upper bound and calculates the cumulative frequencies for each pixel value between the input bounds. This module is called twice by the driver module, once with bounds [0, threshold] and again with bounds [threshold+1, 255]. The cumulative frequency of each pixel value is calculated as defined in \eqref{eq3}.

Cumulative histogram is defined as a recursive algorithm. Hence, we create a 32-bit register \texttt{prev} to store the last calculated value. \texttt{prev} is initialized to 0, for the base case. Also, since this module works within a given bound of [\texttt{idx\_l}, \texttt{idx\_h}], a register \texttt{idx\_offset} is used to store index offset, instead of absolute index. The absolute \texttt{index} is calculated as \texttt{idx\_l + idx\_offset}. The execution does not stop until \texttt{index <= idx\_h} evaluates to \texttt{false}. The input frequency array \texttt{freq}, passes through a multiplexer with \texttt{index} as selector. The selected value passes through an adder, which stores the sum of selected frequency, \texttt{freq[index]}, and previous cumulative frequency, \texttt{prev}, in a separate \texttt{cumu\_freq} array. \texttt{cumu\_freq} is the output of this module.

\subsection{Create\_map()}

\begin{figure*}[ht!]
    \begin{center}
        \includegraphics[width=17.0cm]{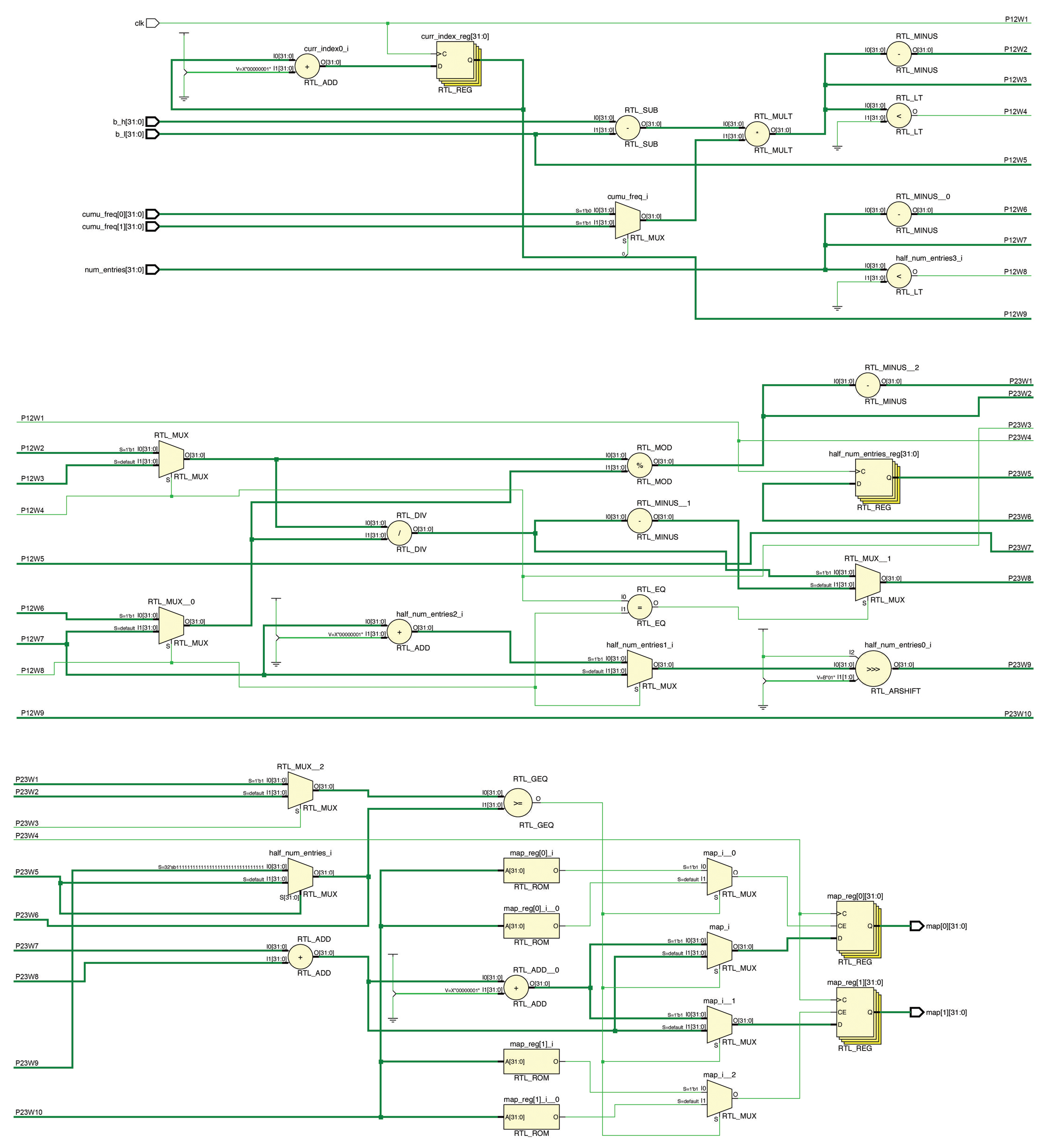}
        \caption{Create\_map() schematic for a binary image with 8 pixels}
        \label{fig6}
    \end{center}
\end{figure*}

This module is the last step in MMBEBHE. It creates a map between the input pixel values and output pixel values. The map is used to create the final output image. It takes the cumulative frequencies, lower bound, and upper bound as input, and outputs the map for the given bounds. Much like Gen\_cumu\_hist(), this module is called twice, once with bounds $[0, threshold]$ and again with bounds $[threshold+1, 255]$. The map is calculated as defined in \eqref{eq4}. Since the output is a pixel value, we use integer division, and round using modulus ($\%$) operator. The output maps are sent forward to be compiled into a single map. 

Calculating map is carried out as described in \eqref{eq4}. This process can be entirely parallelized, but we carry it out serially to reduce hardware size. When the module starts executing, we calculate the half of the total number of pixels, \texttt{num\_entries}, in the input image. This value is calculated as a right shift, i.e., \texttt{num\_entries >> 1}, and is stored in a 32-bit register \texttt{half\_num\_entries}. \texttt{half\_num\_entries} is used to round values up. The input cumulative frequency array \texttt{cumu\_freq} passes through a multiplexer with a \texttt{index} as selector. The selected frequency is use to calculate the corresponding map value. Simultaneously, the remainder with \texttt{num\_entries} is calculated. If the remainder is greater than \texttt{half\_num\_entries}, the map value is increased by one. This process repeats for each pixel value.

\subsection{MMBEBHE()}

\begin{figure*}[ht!]
    \begin{center}
        \includegraphics[height=20cm]{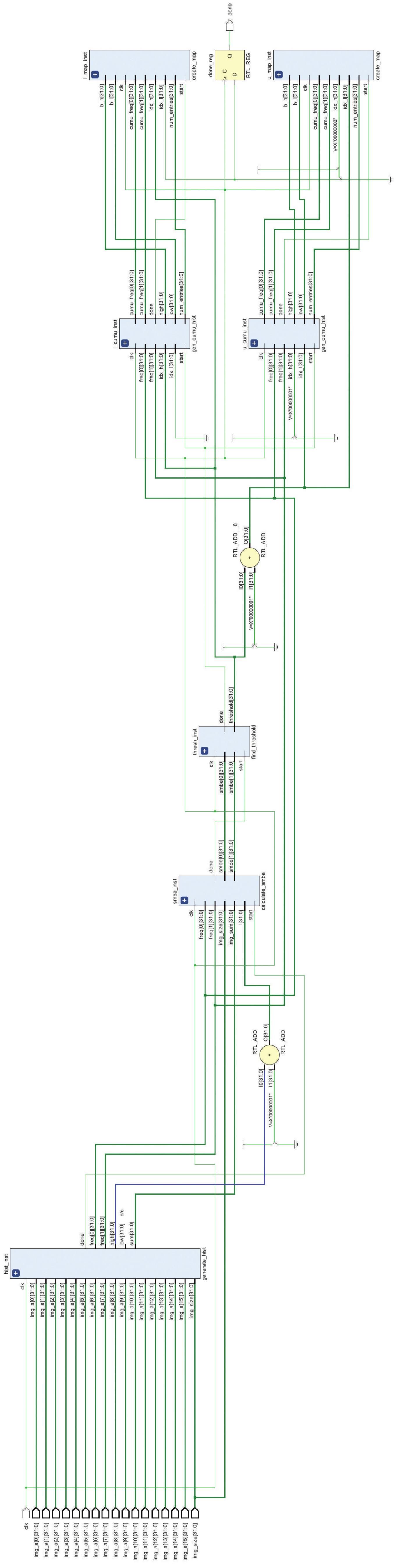}
        \caption{MMBEBHE Schematic for a binary image with 8 pixels}
        \label{fig7}
    \end{center}
\end{figure*}

This is the driver module, responsible for pipelining the other modules. Figure~\ref{fig7} shows how the differed mmodules interact with each other. Our implementation takes image and the image size as input, and outputs a map from input pixel values to output pixel values. To get the final image, the value of each pixel is replaced with corresponding value in the output map.

\section{Experimental Results} 

The map generated by the FPGA on synthesizing our implementation matched the simulation, and result from MATLAB. The output map can be used to recreate the equalized image. Figure~\ref{fig8}, and Figure~\ref{fig9} compare the original image, image created by FPGA, and image created by MATLAB. Results from MATLAB and FPGA are visually and objectively similar, as depicted by their histograms. Therefore, our implementation successfully recreates the results of MMBEBHE on an FPGA, without using floating point arithmetic.

\begin{figure*}[ht!]
    \begin{center}
        \includegraphics[width=14.0cm]{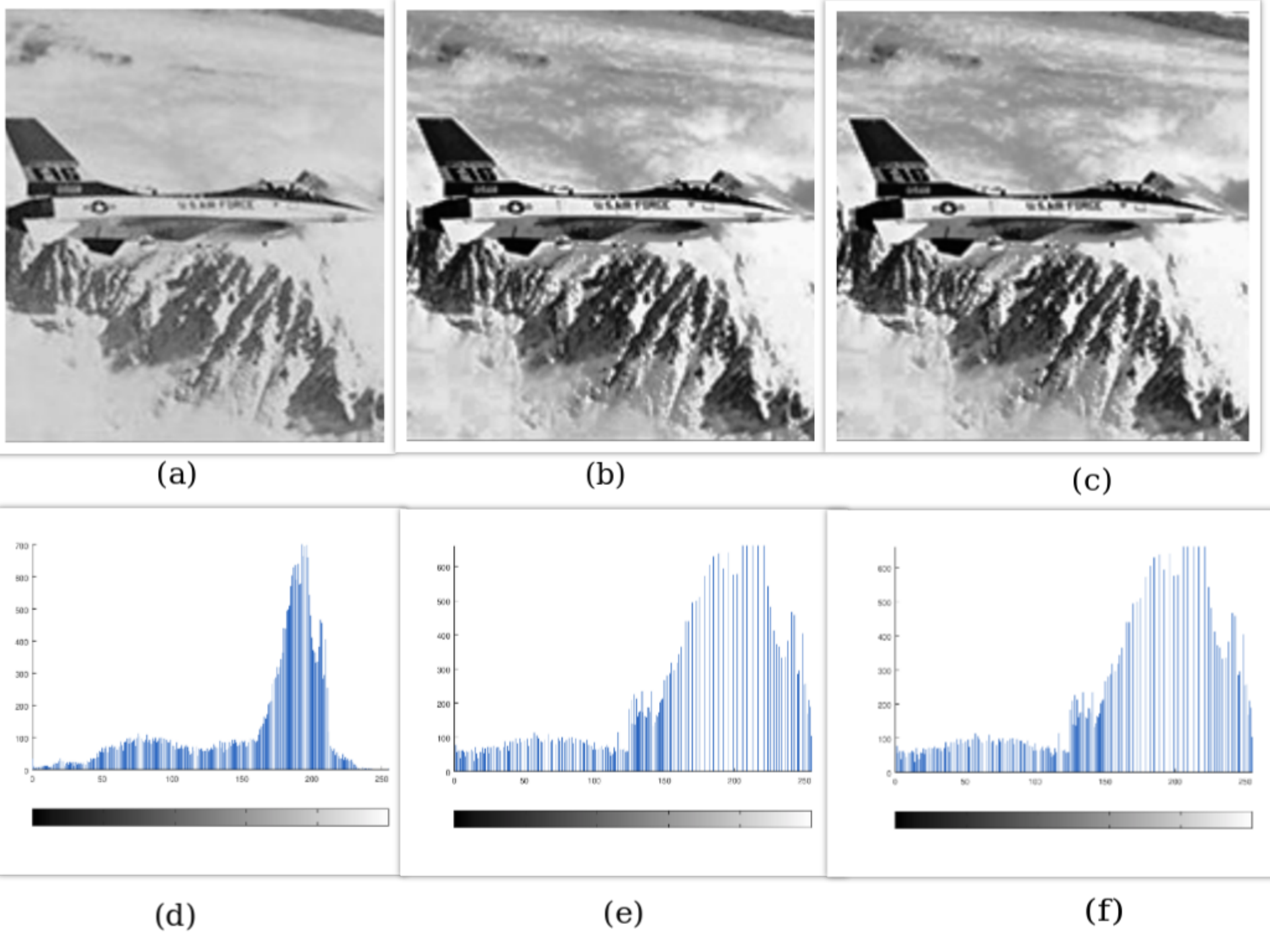}
        \caption{F16 Images Comparison}
        \label{fig8}
        \small{(a): Original image; (b): output from FPGA; (c): Output from MATLAB. (d),(e), \& (f) are histograms of (a),(b), \& (c) respectively.}
    \end{center}
\end{figure*}

\begin{figure*}[ht!]
    \begin{center}
        \includegraphics[width=14.0cm]{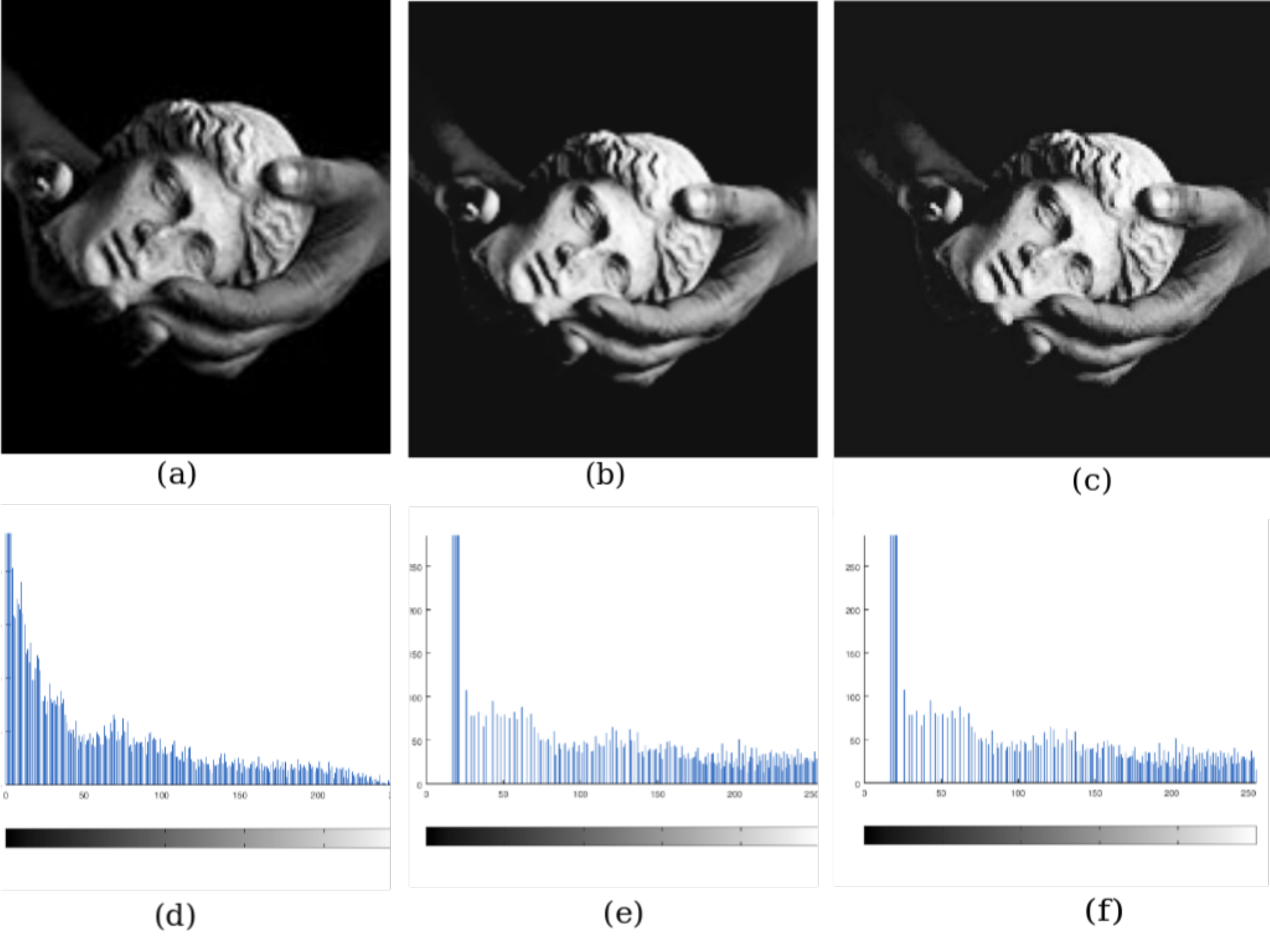}
        \caption{Hands Output Images Comparison}
        \label{fig9}
        \small{(a): Original image; (b): output from FPGA; (c): Output from MATLAB. (d),(e), \& (f) are histograms of (a),(b), \& (c) respectively.}
    \end{center}
\end{figure*}

Although timing actual execution of logic on FPGA is cumbersome, we were able to get an approximate execution time of each logical module through simulations. Table~\ref{tab1} shows comparison between execution times of each logical module in our FPGA implementation with 300 MHz clock, and floating point MATLAB implementation.

\begin{table*}[ht!]
\caption{Timings of MMBEBHE on F16 Image --- FPGA vs MATLAB}
\begin{center}
\begin{tabular}{|l|p{4.5cm}|p{3cm}|}
\hline
\textbf{Module}            & 
\textbf{FPGA Simulation Timing} (300 MHz clock) (in $\mu$s) & 
\textbf{MATLAB} Timing (in $\mu$s) \\
\hline
Generate\_hist() & 
207.68 & 
268                    \\
\hline
Calculate\_SMBE() & 2.57                          & 31                    \\
\hline
Find\_Threshold() & 2.57                          & 12                    \\
\hline
Gen\_Cumu\_Hist() & 2.6                           & \multirow{2}{*}{78}                    \\
\cline{1-2}
Create\_Map()   & 2.6                           &                   \\
\hline
\end{tabular}
\label{tab1}
\end{center}
\end{table*}

Table~\ref{tab3}, and Table~\ref{tab4} show the utilization report of our implementation. Compare this to the resource utilization report of Histogram Equalization as computed by Sawmya and Paily\cite{b9}, in Table~\ref{tab2}.

\begin{table*}[ht!]
\caption{Resource Utilization of Histogram Equalization by Sowmya and Paily\cite{b9}}
\begin{center}
\begin{tabular}{|l|l|}
\hline
Device      & xc2vp30-7ff896                             \\
\hline
I/O Cells   & 32 of 556 (5\%)                            \\
\hline
Block RAMs  & 16 of 136(11\%)                            \\
\hline
Time Period & 5ns                                        \\
\hline
Power       & 148mW         \\
\hline
\end{tabular}
\label{tab2}
\end{center}
\end{table*}

\begin{table*}[ht!]
\caption{DSP Utilization table}
\begin{center}
\begin{tabular}{|l|l|l|l|}
\hline
Site Type              & Used & Available & Util\% \\ \hline
DSPs                   & 6        & 90        & 6.67   \\ \hline

\end{tabular}
\label{tab3}
\end{center}
\end{table*}

\begin{table*}[ht!]
\caption{Slice Logic Utilization table}
\begin{center}
\begin{tabular}{|l|l|l|l|}
\hline
Site Type              & Used & Available & Util\% \\ \hline
1) Slice LUTs            & 6923 &  20800     & 33.28   \\ \hline
\quad a) LUT as logic           & 6315  & 20800     & 30.36   \\ \hline
\quad b) LUT as memory          & 608    & 9600      & 6.33   \\ \hline
2) Slice Registers        & 952   & 41600     & 2.29   \\ \hline
\quad a) Registers as Flip Flop & 929  & 41600     & 2.23   \\ \hline
\quad b) Registers as Latch     & 23   & 41600     & 0.06   \\ \hline
3) F7 Muxes           & 1916 &  16300     & 11.75   \\ \hline
4) F8 Muxes            & 525 &  8150     & 6.44   \\ \hline
\end{tabular}
\label{tab4}
\end{center}
\end{table*}

\section{Conclusions and Future Work}
We present a successful implementation of MMBEBHE on FPGA. We are able to replicate the results of MMBEBHE as found in MATLAB and ModelSim simulations on our FPGA. The future work could potentially include performing the MMBEBHE on larger images, as well as optimizing it to increase parallel processing and hardware concurrency.


\begin{thebibliography}{99}
\bibitem{b1}  J. Zimmerman, S. Pizer, E. Staab, E. Perry, W. McCartney, B. Brenton, ``Evaluation of the effectiveness of adaptive histogram equalization for contrast enhancement,''  IEEE Trans. on Medical Imaging, pp. 304-312, Dec. 1988.

\bibitem{b2} Y. Li, W. Wang, D. Y. Yu, ``Application of adaptive histogram equalization to x-ray chest image,''  Proc. of the SPIE, vol. 2321, pp. 513-514, 1994.

\bibitem{b3} Yeong-Taeg Kim, ``Method and circuit for video enhancement based on the mean separate histogram equalization,''filed in a Korean patent, March 9, 1996, Appl. No. 6219.

\bibitem{b4} Y. T. Kim, ``Contrast enhancement using brightness preserving bi-histogram equalization,''  IEEE Trans. Consum. Electron., vol. 43, no. 1, pp. 1–8, Feb. 1997. 

\bibitem{b5} Chen and A. Ramli, ``Minimum mean brightness error bi- histogram equalization in contrast enhancement,''  IEEE Trans. Consum.Electron., pp. 1310–1319 Nov. 2003.

\bibitem{b6} A. Trost, B. Zajc Zemva, "Pogrammable System for Image Processing" in Field-Programmable Logic and Applications, Elsevier, pp. 490-494, 1998.

\bibitem{b7} Wang Bing-jian, Liu Shang-qian, Qing Li, Zhou Hui-xin, "A realtime contrast enhancement algorithm for infrared images based on plateau histogram", Infrared Physics and Technology Elsevier, pp. 77-82, 2006.

\bibitem{b8} Abduallah M. Alsuwailem, Saleh A. Alshebeili, "A New approach for real-time histogram equalization using FPGA", Proceedings of International Symposium on Intelligent Signal Processing and Communication Systems, 2005.

\bibitem{b9} S. Sowmya and R. Paily, ``FPGA implementation of image enhancement algorithms,'' 2011 International Conference on Communications and Signal Processing, Calicut, 2011, pp. 584-588.


\end{thebibliography}
\end{document}